\documentclass[letterpaper, 10 pt, conference]{ieeeconf}  

\IEEEoverridecommandlockouts                              

\usepackage{graphics} 
\usepackage{graphicx}
\usepackage{xcolor}

\usepackage{booktabs} 
\usepackage{array} 

\usepackage{algorithmic}
\usepackage[linesnumbered,ruled,vlined]{algorithm2e}

\makeatletter
\let\NAT@parse\undefined
\makeatother

\usepackage{url}
\usepackage[hidelinks]{hyperref}
\usepackage{amsmath, xparse}

\title{\LARGE \bf
    Extended Friction Models for the Physics Simulation of Servo Actuators
}

\author{Marc Duclusaud$^{1*}$, Grégoire Passault$^{1*}$, Vincent Padois$^{2}$, Olivier Ly$^{1}$
\thanks{$^{1}$Univ. Bordeaux, CNRS, LaBRI, UMR 5800, 33400 Talence, France. Corresponding author: Marc Duclusaud, e-mail: \texttt{marc.duclusaud@u-bordeaux.fr}}
\thanks{$^{2}$Inria, Auctus, 33400 Talence, France.}
\thanks{$^{*}$Both authors contributed equally to this work.
\newline This study has received financial support from the French government in the framework of the France 2030 program, Initiative of Excellence (IdEx) University of Bordeaux / RRI ROBSYS.}
}

\begin{document}

\maketitle
\thispagestyle{empty}
\pagestyle{empty}

\begin{abstract}

Accurate physical simulation is crucial for the development and validation of control algorithms 
in robotic systems. Recent works in Reinforcement Learning (RL) take notably advantage of extensive 
simulations to produce efficient robot control. State-of-the-art servo actuator models generally fail 
at capturing the complex friction dynamics of these systems. This limits the transferability
of simulated behaviors to real-world applications. In this work, we present 
extended friction models that allow to more accurately simulate servo actuator dynamics. We propose a comprehensive 
analysis of various friction models, present a method for identifying model parameters using recorded 
trajectories from a pendulum test bench, and demonstrate how these models can be integrated into physics engines.
The proposed friction models are validated on four distinct servo actuators and tested on 2R manipulators, 
showing significant improvements in accuracy over the standard Coulomb-Viscous model. Our results highlight 
the importance of considering advanced friction effects in the simulation of servo actuators to 
enhance the realism and reliability of robotic simulations.

\end{abstract}

\section{Introduction}

Physical simulation is a crucial element in the development of robotic systems, allowing for the testing 
and validation of control algorithms prior to real-world deployment. This is particularly vital in 
Reinforcement Learning (RL), which has seen significant advancements in robotics in recent years due to 
its capability to implement robust, versatile, and adaptive behaviors~\cite{rlRoboticsSurvey2013}. 
Leveraging the computational power of modern hardware, RL algorithms can learn control policies through trial 
and error interactions with a simulated physical environment~\cite{todorov2012mujoco}~\cite{isaacGym}, where iterations 
can be performed safely and at a much higher rate than in the real world. In this context, the accuracy of
the simulation is paramount to ensure the transferability of the learned policies to the real-world system.

A common approach to ensuring consistency between simulation and reality in robotics is to identify the dynamics of the 
actuators~\cite{zhu2021survey}. Several studies have proposed identifying servo actuator dynamics by recording 
trajectories on the real system and minimizing the error between simulated and recorded trajectories. This process
involves tuning the servo actuator model parameters using methods like iterative coordinate descent~\cite{openAI2020dexterous} 
or genetic algorithms~\cite{fabre2017dynaban}\cite{masuda2023sim}. While these methods have demonstrated promising results, they are 
constrained by the complexity of servo actuator dynamics, which can be challenging to accurately capture 
using simple models.

\begin{figure}[t]
    \includegraphics[width=1.\columnwidth]{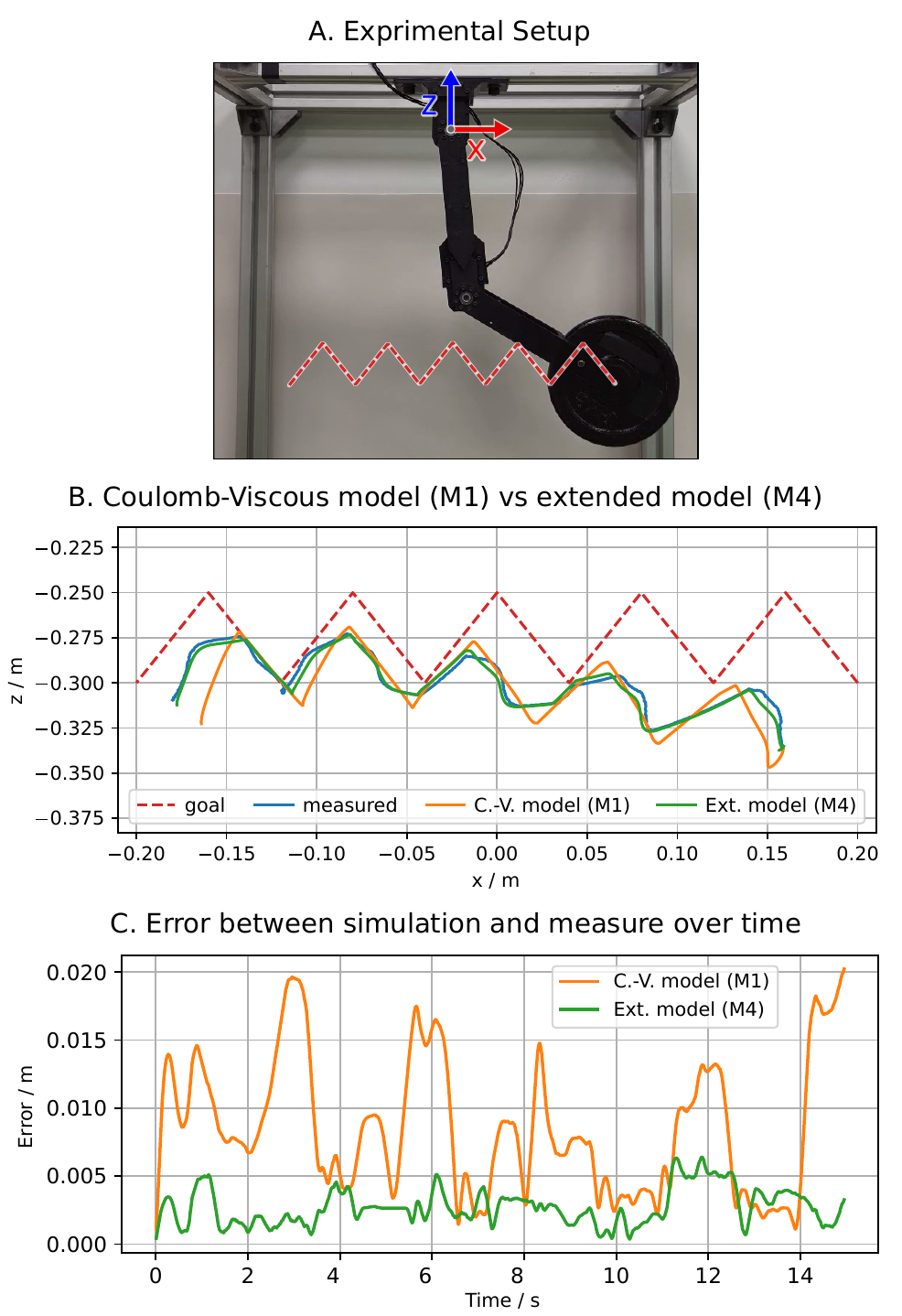}
    \vspace{-1.75em}
    \caption{Comparison on a 2R arm of the Coulomb-Viscous model ($\mathcal{M}_1$), classically used in physics 
    simulators, and a proposed servo actuator model ($\mathcal{M}_4$). The 2R arm (\textbf{A.}) is composed of Dynamixel 
    MX-106 and MX-64 controlled with low gains to mimic the control modes typically used in RL 
    applications. A triangular wave path is tracked by the real system and simulated using two different friction models. 
    The simulated and measured trajectories are presented in (\textbf{B.}), and the error 
    between simulation and measure over time in (\textbf{C.}), highlighting the importance of accounting for various friction 
    effects in the simulation of servo actuators.}
    \label{fig:catch-eye}
    \vspace{-1.75em}
\end{figure}

To capture the complex dynamics of servo actuators, several studies have proposed using supervised machine learning techniques. 
For example, Lee et al. introduced a multi-layer perceptron that uses a history of position errors and velocities as 
input to predict the torque applied by the servo actuator~\cite{actuatorNet2019}. Although their network outperformed an 
ideal servo actuator model with infinite bandwidth and zero latency, the lack of comparison with an identified model makes 
it challenging to fully assess its performance. Another study by Serifi et al. proposed a transformer-based augmentation 
to correct the simulated position of servo actuators~\cite{transformerDisney2023}. While this approach showed promising results, 
it does not offer a physical interpretation for the proposed corrections.

In this work, we focus on simulating servo actuators using models that accurately capture the 
effects of friction. Friction is a complex phenomenon that can, in a lubricated environment such 
as a motor gearbox, be decomposed into two main components: \textbf{static friction} (also known 
as dry friction) and \textbf{viscous friction}~\cite{albender2008characterization}. Static friction is 
the force that opposes motion when the joint is stationary.
It represents a threshold force that must be overcome to initiate motion. 
Once the joint starts moving, static friction decreases, and viscous friction increases in a 
pre-sliding regime. At higher speeds, viscous friction becomes dominant over static friction, 
leading to a sliding regime.

Viscous friction is typically modeled as a linear function of velocity, acting as a damper. 
Static friction, on the other hand, is often modeled as a constant force, a concept originally 
described by Coulomb in the 18th century~\cite{coulomb1785theorie}. Despite the identification of 
other effects impacting static friction, such as the Stribeck effect~\cite{pennestri2016review} and load 
dependence~\cite{bittencourt2012static}, the Coulomb-Viscous model remains widely used in popular physics simulators 
like MuJoCo~\cite{todorov2012mujoco} and Isaac Gym~\cite{isaacGym}. Consequently, simulations 
of servo actuators in these environments may not accurately reflect real-world behaviors, 
particularly when controlled with low gains, which is common in RL applications.
An enhanced friction model was proposed in \cite{masuda2023sim} to account for some of these effects, 
but it does not adequately quantify their individual contributions to the simulation improvement.

To address these limitations, we propose servo actuator models exploiting extended friction models. 
The main contributions of this work are as follows:
\begin{itemize} 
    \item \textbf{Friction Model Analysis}: A comprehensive analysis of the different friction models and their effects on the dynamics of servo actuators. 
    \item \textbf{Parameter Identification Method}: A method to identify the parameters of these models using recorded trajectories from a pendulum test bench. 
    \item \textbf{Simulation Integration}: A detailed explanation on how to simulate the dynamics of servo actuators in a physics engine using the proposed models. 
\end{itemize}

In the following sections, we present the background and motivation for this work, introduce the extended friction models,
explain how to simulate servo actuators with these models, and present the results of our experiments.
To demonstrate the effectiveness of the proposed models, four distinct servo actuators are identified: 
Dynamixel MX-64~\cite{mx64} and MX-106\cite{mx106}, eRob80:50 and eRob80:100, which are both eRob80\cite{erob80} 
with respectively 1:50 and 1:100 reduction ratios. We then compare the Coulomb-Viscous model
with the extended models on two 2R arms. Such a comparison is presented in 
Fig.~\ref{fig:catch-eye}, highlighting the importance of accounting for different 
friction effects in the simulation of servo actuators. Every experiment conducted in this work is presented in 
the video accompanying this paper\footnote[1]{\href{https://youtu.be/5XPEEKDnQEM}{https://youtu.be/5XPEEKDnQEM}}.

\section{Background and motivation}

This section provides an overview of the complexity involved in simulating 
friction in servo actuators and outlines the motivation for this work.

\subsection{Test bench dynamics}
\label{subsec:dynamics}

We consider a servo actuator on a pendulum test bench composed a rigid link of length $l$ and a point load of mass $m$
subjected to gravity acceleration $g$. In this configuration, presented in Fig. \ref{fig:drive_backdrive}, 
the dynamics of the system can be expressed as
\begin{equation}
    \tau_m + \tau_e (\theta) + \tau_f = J \ddot \theta,
    \label{eq:dynamics}
\end{equation}

where $\theta$ is the servo actuator position, $\tau_m$ the motor torque, 
$\tau_e (\theta) = -m g l \sin(\theta)$ the external torque caused
by gravity, $\tau_f$ the torque caused by friction and $J$ the inertia.
The inertia $J$ is the sum of the load inertia $ml^2$ and the servo actuator 
apparent inertia $J_m$ (sometimes referred to as \textit{armature}).
When the rotor inertia $J_r$ is known from constructor data, $J_m$ is given by $J_m = N^2 J_r$ with $N$ the inverse 
of the gear ratio~\cite{lynch2017modern}. Otherwise, $J_m$ is an extra parameter to identify.
In the following notations, we drop the dependency of $\tau_e$ to $\theta$ for clarity.

\subsection{Friction model}
\label{subsec:friction}

Friction can be seen as a force preventing the motion of the system.
In rotational systems, it results in a torque $\tau_f$ opposing the direction of the motion.
The most commonly used model is the Coulomb-Viscous model~\cite{pennestri2016review}, generally expressed as
\begin{equation}
    \tau_f = - K_v \dot \theta - sign(\dot \theta) K_c,
\end{equation}

where $K_c$ and $K_v$ are two positive parameters representing respectively 
the Coulomb friction, which must be overcome for the system to 
start moving, and the viscous friction, which is countering the 
velocity and acts as a damper.
With this model, $\tau_f$ is a passive torque countering the direction of the motion.
This formulation implies zero friction when $\dot \theta = 0$, which does not reflect 
reality. Moreover, the friction becomes discontinuous when the direction of motion changes, 
which is both physically inaccurate and unsuitable for simulation.

In this work, we define friction as a torque budget $\tau_f^m$ available to prevent the motion.
For the Coulomb-Viscous model, we express it as
\begin{equation}
    \mathcal{M}_1: \tau_f^m = K_v | \dot \theta | + K_c
    \label{eq:coulomb}
\end{equation}

In a simulator using time steps of $\Delta t$ seconds, 
stopping the motion can be interpreted as bringing the
velocity to zero at the next time step, leading to the condition: 
$\dot \theta + \ddot \theta \Delta t = 0$.\\
\noindent Based on equation (\ref{eq:dynamics}), assuming friction is the stopping factor, 
this would require a torque $\tau_{f}^{stop}$ of:
\begin{equation}
    \tau_{f}^{stop} = - \biggl(\frac{J}{\Delta t} \dot \theta + \tau_{m} + \tau_{e}\biggr)
    \label{eq:stop}
\end{equation}

Thus, the resistive torque $\tau_f$ caused by friction is given by 
$\tau_{f}^{stop}$, as computed in equation (\ref{eq:stop}), 
and  limited to the range $[-\tau_f^m, \tau_f^m]$:
\begin{equation}
    \tau_{f} = clip(\tau_{f}^{stop}, [-\tau_{f}^m, \tau_{f}^m])
\end{equation}


\subsection{Drive and backdrive torques}

\begin{figure}[!ht]
    \vspace*{-1.5em}
    \centering
    \includegraphics[width=.85\columnwidth]{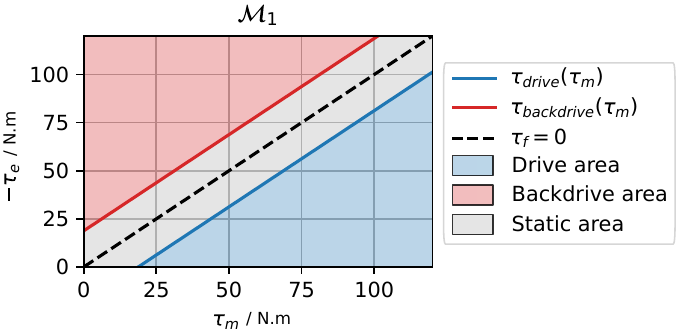}
    \vspace{-.5em}
    \caption{
        Drive/backdrive diagram for the Coulomb-Viscous model. The static area (gray)
        is where the system verify $\tau_m + \tau_e + \tau_f = 0$. In the drive (blue) 
        and backdrive (red) areas $|\tau_f| = \tau_f^m$. On the blue and red lines, $\tau_f^m=|\tau_e+\tau_m|$. 
        The dashed line corresponds to configurations where 
        the system is at equilibrium without the need for friction to act ($\tau_m = -\tau_e$).
    }
    \label{fig:coulomb-drive}
    \par
    \vspace{-.5em}
\end{figure}

Considering a system where the motor torque $\tau_m$ and the external torque $\tau_e$
are opposing each other, 3 different states are possible: the servo actuator is moving
in the direction of $\tau_m$ (drive state), the system is in a static equilibrium, or the servo actuator
is moving in the direction of $\tau_e$ (backdrive state). These states can be represented on drive/backdrive diagrams,
as presented in~\cite{wilfrido2021load}\cite{zhu2019design}. 
Fig. \ref{fig:coulomb-drive} presents such a diagram for the Coulomb-Viscous model 
presented in equation (\ref{eq:coulomb}). \\

For a given $\tau_m$, one calls \textbf{drive torque}, denoted it by $\tau_{drive}(\tau_m)$ 
the highest value of $\tau_e$ keeping $(\tau_m, \tau_e)$ in the Drive Area, and \textbf{backdrive torque}, 
denoted by $\tau_{backdrive}(\tau_m)$, the smallest value of $\tau_e$ keeping $(\tau_m, \tau_e)$ in the Backdrive Area.

\begin{figure}[!ht]
    \includegraphics[width=1.\columnwidth]{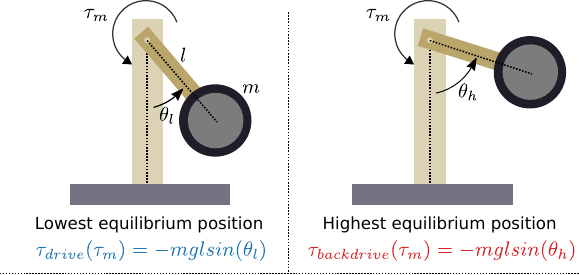}
    \includegraphics[width=1.\columnwidth]{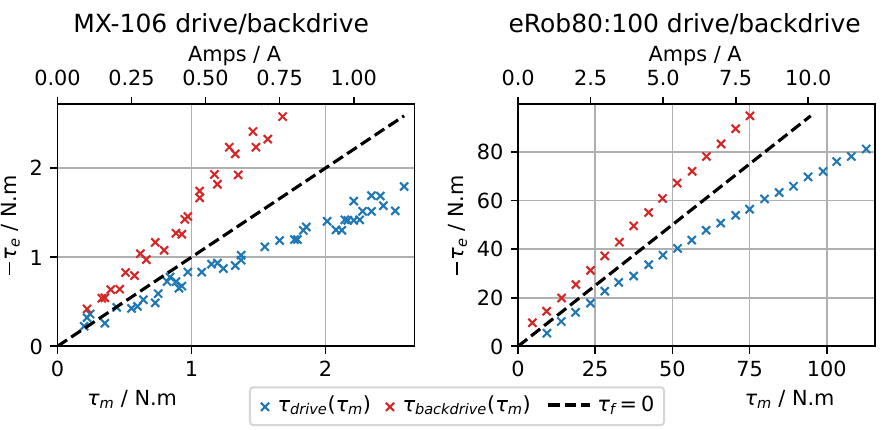}
    \par
    \vspace{-.5em}
    \caption{
        On the top, the experiment used to find drive and backdrive torques.
        Below, the drive/backdrive diagrams obtained for the MX-106 (bottom left) and
        eRob80:100 (bottom right) servo actuators. 
    }
    \label{fig:drive_backdrive}
    \par
    \vspace{-.5em}
\end{figure}

For a fixed $\tau_m$, the values $\tau_{drive}(\tau_m)$ and $\tau_{backdrive}(\tau_m)$ can be identified on a pendulum test bench.
From a static equilibrium position, such as presented in Fig. \ref{fig:drive_backdrive}, 
the system can be moved by hand to the extremum positions of the static area.
In the lowest position $\tau_e = \tau_{drive}(\tau_m)$ and in the highest 
$\tau_e = \tau_{backdrive}(\tau_m)$. 

The drive/backdrive diagrams of the MX-106 and eRob80:100 servo actuators presented in Fig. 
\ref{fig:drive_backdrive} show the limits of the Coulomb-Viscous model. Notably the drive and backdrive
torques show that the friction is increased by the involved torques: it is said 
to be load-dependent. 

\section{Extended friction models}
\label{sec:friction-models}

The Coulomb-Viscous model presented in the previous section is typically used in physics simulation.
Based on Fig. \ref{fig:drive_backdrive}, we can expect this type of model to lack expressiveness to simulate servo actuators.
In this section, we describe those effects and provide extended models to account for them.
All parameters introduced in this section are positive.

\subsection{Stribeck model (5 parameters)}

The static friction is known to be higher when the system is not moving due to the imbrication of
the asperities of the surfaces in contact, and lower when the system is moving~\cite{albender2008characterization}.
This effect, known as the Stribeck effect, can be modeled as an exponential decay of the friction, ensuring
continuity and a smooth transition between the two regimes:
\begin{equation}
    \mathcal{M}_2:
    \tau_f^m
    =
    K_v | \dot \theta |
    +
    K_c
    +
    e^{-\left|\frac{\dot \theta}{v_s}\right|^\alpha}
    K_c^s
\end{equation}

The parameter $K_c^s$ is the Stribeck-Coulomb friction, which is an extra static friction when not moving.
The Stribeck velocity $v_s$ parametrizes the zone of influence of the Stribeck effect and $\alpha$ its curvature.
This model is used with Dynamixel servo actuators in \cite{fabre2017dynaban}.

\subsection{Load-dependent model (3 parameters)}

The effect of static friction can be augmented by the load exerted on the gearbox.
Intuitively, gear teeth have to slide against each other in order to get the system moving.
This sliding is made harder by the applied load $|\tau_m - \tau_e|$:
\begin{equation}
    \mathcal{M}_3:
    \tau_f^m
    =
    K_v | \dot \theta |
    +
    K_c
    +
    K_l
    |\tau_m - \tau_e|
\end{equation}

The introduced parameter $K_l$ is the load-dependent friction, which adds extra friction proportionally to the load.

\subsection{Stribeck load-dependent model (7 parameters)}

Extra static friction produced by load-dependence is also known to be reduced when the system is moving.
By combining the previous models, a Stribeck load-dependent 
model~\cite{bittencourt2012static}\cite{mori2023identification} can be formulated as
\begin{equation}
    \begin{aligned}
    \mathcal{M}_4:
    \tau_f^m
    = K_v | \dot \theta |
    + K_c + K_l |\tau_m - \tau_e|
    \\
    + 
    e^{-\left|\frac{\dot \theta}{v_s}\right|^\alpha}
    [K_c^s + K_l^s |\tau_m - \tau_e|]
    \end{aligned}
\end{equation}

The load-dependence is now represented by two terms, $K_l^s$ being the Stribeck-load-dependent friction,
which is an extra load-dependent friction for being at rest.

\subsection{Directional model (9 parameters)}

As suggested by \cite{wang2015directional}, the efficiency of the gearbox can be directional.
Some systems, like endless screws used in linear actuators, are even not backdriveable at all
(they are said to be \textit{self-locking}).
To account for this in the friction model, the load-dependent parameters can be split up
depending on the direction:
\begin{equation}
    \begin{aligned}
    \mathcal{M}_5:
    \tau_f^m
    = K_v | \dot \theta |
    + K_c + |K_m \tau_m - K_e \tau_e|
    \\
    + 
    e^{-\left|\frac{\dot \theta}{v_s}\right|^\alpha}
    [K_c^s + |K_m^s \tau_m - K_e^s \tau_e|]
    \end{aligned}
\end{equation}

In this model, $K_m$ and $K_e$ are the motor and external load-dependent frictions, and 
$K_m^s$ and $K_e^s$ are the motor and external Stribeck-load-dependent frictions.
In the case of a self-locking endless screw, we would have $K_e^s > 1$.

\subsection{Quadratic model (11 parameters)}

Finally, it is shown \textit{e.g.} by~\cite{wilfrido2021load} that the effect of friction can be
quadratic in the load in the case of harmonic drives. This effect, which can be observed on 
the eRob:100 drive/backdrive diagram in Fig. \ref{fig:drive_backdrive}, can be modeled as
\begin{equation}
    \begin{aligned}
    \mathcal{M}_6:
    \tau_f^m
    = K_v | \dot \theta |
    + K_c + |K_m \tau_m - K_e \tau_e|
    \\
    + 
    e^{-\left|\frac{\dot \theta}{v_s}\right|^\alpha}
    [K_c^s + |K_m^s \tau_m - K_e^s \tau_e| + Q] \\
    Q = 
    \left\{
        \begin{aligned}
            & K_e^q \tau_e^2 \ \ & if \ \ |\tau_m| > |\tau_e| \\
            & K_m^q \tau_m^2 \ \ & if \ \ |\tau_m| < |\tau_e|
        \end{aligned}
    \right.
    \end{aligned}
\end{equation}

The parameters $K_m^q$ and $K_e^q$ are the motor and external load-dependent quadratic frictions.

\subsection{Drive/backdrive diagrams}

\begin{figure}[!ht]
    \includegraphics[width=1.\columnwidth]{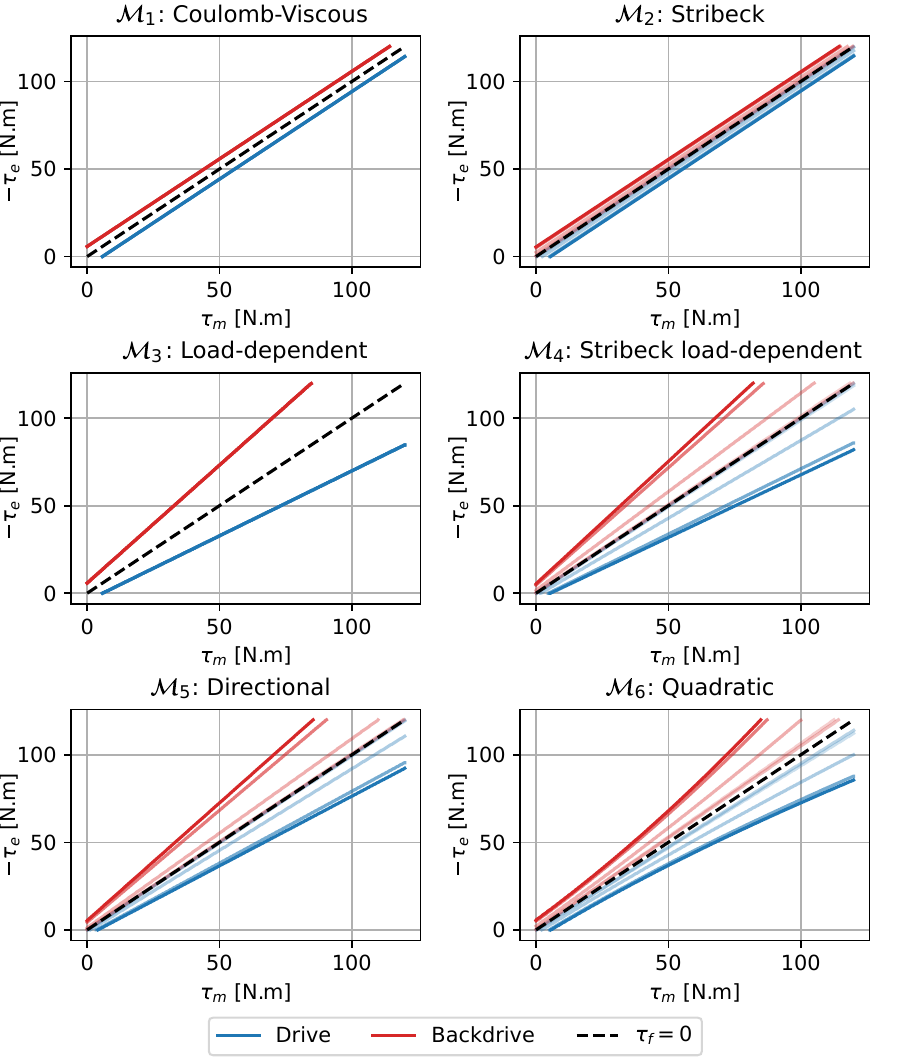}
    \caption{
        Drive/backdrive diagrams for the different proposed models, with optimal parameters
        fitted during the eRob80:100 servo actuator identification. The lines with lower opacity denotes 
        the effect of velocity (1 rad/s per step).
    }
    \label{fig:db_models}
\end{figure}

The drive/backdrive diagrams for the proposed models are presented in Fig. \ref{fig:db_models}.
The parameters are fitted for the eRob80:100 servo actuator using the method presented in section 
\ref{sec:identification}. This figure illustrates the impact of the different effects taken into account on the
drive and backdrive curves.

\section{Simulating servo actuators with friction}

In this section, a detailed approach is presented for accurately simulating the behavior of a 
position controlled servo actuator, incorporating the friction models introduced in the previous section.

\subsection{Servo actuator model}
\label{subsec:sim-servo}

A servo actuator model $\mathcal{S}$ is a function that takes the current state of the system $(\theta, \dot \theta)$
and a target position $\theta^d$, and outputs a torque $\tau_m$ to apply to the system. 
\begin{equation}
    \tau_m = \mathcal{S}(\theta, \dot \theta, \theta^d)
\end{equation}

The servo actuator model is composed of two parts: the control law and the motor model.
To simulate the servo actuator, the control law should be known. Two exemple of standard control laws 
are presented hereafter.

\subsection{Voltage control law}

A voltage control law computes the 
voltage $U$ to apply to the motor from the current state and the target position. It generally involves a 
proportional-integral-derivative (PID) controller, whose gains can be given by the constructor or identified. 
The control law can be written as
\begin{equation}
    U = clip(PID(\theta, \dot \theta, \theta^d), [-U_{max}, U_{max}]), 
\end{equation}

where $U_{max}$ is the motor maximum voltage and $clip$ a function that bounds a value to a given interval.
The motor torque can then be derived from DC motor equations \cite{lynch2017modern}:
\begin{equation}
    \tau_m = \frac{k_t}{R} U - \frac{k_t^2}{R} \dot \theta,
    \label{eq:voltage-dc-sim}
\end{equation}

where $R$ is the internal motor resistor and $k_t$ its torque constant. 
If they are not known, $k_t$ and $R$ are extra parameters to identify.
In this work the motor model refers to the motor and the reducer. Thus, the torque 
constant $k_t$ is the product of the motor torque constant and the reducer ratio.

Equation (\ref{eq:voltage-dc-sim}) assumes the use of a \textit{drive/brake} controller, which is the most common.
In the case of a \textit{drive/coast} controller, different equations can be derived, involving an electric
time constant which depends on the motor inductance $L$.
See \cite{sihite2019derivation} for a comprehensive derivation.

\subsection{Current control law}

A current control law computes a target current $I$ to apply to the motor from the current state and the target position.
Ensuring the target current to flow in the motor is itself ensured by another high frequency electronics loop.
It can be written as
\begin{equation}
    I = clip(PID (\theta, \dot \theta, \theta^d), [I_{min}, I_{max}]),
\end{equation}
\begin{equation*}
    \text{with}
    \hspace{0.5cm}
    \begin{aligned}
        I_{min} = & \max(-I_{emf}, - I_{heat}), \\
        I_{max} = & \min(I_{emf}, I_{heat}),
    \end{aligned}
\end{equation*}

where $I_{emf} = \frac{1}{R} [U_{max} - k_t \dot \theta]$ is the maximum current allowable by the
input voltage, and $I_{heat}$ a limit current to dissipate heat.
The motor torque can then be derived from the current using the torque constant:
\begin{equation}
    \tau_m = k_t I
\end{equation}

\subsection{Simulating the pendulum test bench}

Using the dynamics of the test bench (section \ref{subsec:dynamics}), one of the forementionned 
friction models (section \ref{subsec:friction} and \ref{sec:friction-models}) and a servo actuator model 
(section \ref{subsec:sim-servo}), the simulation of a pendulum test bench can be done
using Algorithm \ref{alg:simulation}.

\begin{algorithm}
    \KwIn{
        $\Delta t$:~timestep, \\
        \hspace{1.1cm} $(\theta_k, \dot \theta_k)$:~state at step $k$, \\
        \hspace{1.1cm} $\theta^d_{k+1}$:~desired position at step $k+1$, \\
        \hspace{1.1cm} $\mathcal{M}$:~friction model (section~\ref{sec:friction-models}), \\
        \hspace{1.1cm} $\mathcal{S}$:~servo actuator model (section~\ref{subsec:sim-servo})
    }
    \KwOut{
        $(\theta_{k+1}, \dot \theta_{k+1})$:~state at step $k+1$
    }

        $\tau_{m,k} \leftarrow \mathcal{S}(\theta_k, \dot \theta_k, \theta^d_{k+1})$; servo actuator model\\
        $\tau_{e,k} \leftarrow -m g l sin(\theta_k)$; test bench dynamics\\
        $J \leftarrow ml^2 + J_m$; test bench dynamics\\
        $\tau_{f,k}^m \leftarrow \mathcal{M}(\tau_{m,k}, \tau_{e,k}, \dot \theta_k)$; friction model\\
        $\tau_{f,k}^{stop} \leftarrow - (\frac{J}{\Delta t} \dot \theta_k + \tau_{m,k} + \tau_{e,k})$; torque to stop\\
        $\tau_{f,k} \leftarrow clip(\tau_{f,k}^{stop}, [-\tau_{f,k}^m, \tau_{f,k}^m])$; applied friction\\
        $\ddot \theta_k \leftarrow \frac{\tau_{m,k} + \tau_{e,k} + \tau_{f,k}}{J}$; angular acceleration\\
        $\dot \theta_{k+1} \leftarrow \dot \theta_k + \ddot \theta_k \Delta t$; integrating dynamics\\
        $\theta_{k+1} \leftarrow \theta_k + \dot \theta_k \Delta t + \ddot \theta_k \frac{1}{2} \Delta t^2$; integrating dynamics

    \caption{Simulating one step of the pendulum dynamics with friction.}
    \label{alg:simulation}
\end{algorithm}

\subsection{Simulation using a physics engine}
\label{subsec:sim-engine}

To simulate a more complex system, physics engines such as MuJoCo~\cite{todorov2012mujoco} 
or Isaac Gym~\cite{isaacGym} can be used.
Such simulators implement the Coulomb-Viscous friction model, which approximates the friction
by a constant static friction and a linear viscous friction.
Because of other considerations such as contacts, equality constraints or multibody dynamics, 
this computation is more convoluted and beyond the scope of this paper.
However, it is similar to the intuition provided in section \ref{subsec:friction}.

To take advantage of this computation, the static and viscous frictions can be updated 
on-the-fly based on the current state of the system using an extended friction model. 
Note that $\tau_e$ and $\tau_f$ are computed simultaneously when solving the dynamics 
optimization problem, making $\tau_e$ unavailable for computing $\tau_f^m$. To address 
this, the previous value of $\tau_e$ can be used to compute the current $\tau_f^m$. 
The fact that the external torque does not vary abruptly in the simulation due to the 
soft handling of constraints makes this approximation valid.

\section{Identification method}
\label{sec:identification}

In this section, we present a method to identify the parameters of the friction and servo actuator models.
The method consist in recording trajectories on a physical test bench as well as in a simulation of 
the corresponding system and then optimize the actuators model parameters to minimize the simulation 
errors with respect to the real trajectories.

\subsection{Trajectories}
\label{subsec:trajectories}

The four following trajectories are recorded using different combinations of masses,
pendulum lengths and servo actuator control laws:

\begin{enumerate}
    \item \textbf{Accelerated oscillations}: Oscillations of constant amplitude with progressively increasing 
    frequency, showing phase and amplitude shifts of the controller.
    \item \textbf{Slow oscillations with smaller sub-oscillations}: Sum of large and slow oscillations
    with smaller oscillations having higher frequency. 
    \item \textbf{Raising and lowering slowly}: The mass is raised and slowly lowered, typically resulting in 
    a plateau where the system remains static until backdriven.
    \item \textbf{Lift and drop}: The mass is lifted, and then released. For a servo actuator using a current control
    law, the current is then set to zero. For a servo actuator using a voltage control law, the H-bridge is released. 
    This ensures the motor electromotive force \cite{lynch2017modern} is absent, distinguishing it from viscous friction.
\end{enumerate}

\subsection{Optimization}

The parameters of the friction and servo actuators models are estimated 
using the CMA-ES genetic algorithm~\cite{hansen2001completelyCMAES}, 
implemented via the \textit{optuna} Python module~\cite{akiba2019optuna}. 
Algorithm \ref{alg:simulation} is used to simulate the trajectories 
previously introduced, and the Mean Absolute Error (MAE) between the 
simulated and measured trajectories serves as the cost function 
for the optimization.

The logs are divided into two sets: 75\% for parameter identification 
and 25\% for validation. The identification process for each model 
was repeated three times, consistently converging to the same 
scores and parameters in under 5 minutes and approximately 
4000 iterations.



\section{Experiments and results}

In this section, we present the protocols and results of the identification and simulation 
of the proposed models on four different servo actuators.

\subsection{Identification}

An identification is performed on MX-64, MX-106, eRob80:50 and eRob80:100 servo actuators, following
the protocol described in section \ref{sec:identification}. For each model from $\mathcal{M}_1$ to $\mathcal{M}_6$, the parameters
of the friction model, the electrical parameters $k_t$ and $R$, and the apparent inertia of the motor $J_m$ are identified.

The Dynamixel servo actuators use standard spur gear reducers, 
while the eRob80 utilize harmonic drives. All servo actuators are position-controlled: Dynamixel using the 
manufacturer's voltage control law, and eRob80 with a custom current control law. 
The trajectories presented in section \ref{subsec:trajectories} are recorded for the 
configurations presented in Table \ref{table:config}, resulting 
in around 100 logs of 6 seconds for each servo actuator. 

\begin{table}[h!]
    \centering
    \caption{Configurations used for identification }
    \resizebox{0.95\columnwidth}{!}{
    \begin{tabular}{lcc}
        \toprule
        \textbf{Variable} & \textbf{Dynamixel} & \textbf{eRob80} \\ 
        \midrule
        Load mass / kg & 0.5 / 1 / 1.5 & 3.1 / 8.2 / 12.7 / 14.6 / 19.6 \\ 
        Pendulum length / m & 0.1 / 0.15 / 0.2 & 0.5 \\ 
        Proportional gain & 4 / 8 / 16 / 32 & 10 / 25 / 50 / 100 \\ 
        \bottomrule
    \end{tabular}
    }
    \label{table:config}
\end{table}

\begin{figure}[!ht]
    \vspace*{1.5em}
    \includegraphics[width=1.\columnwidth]{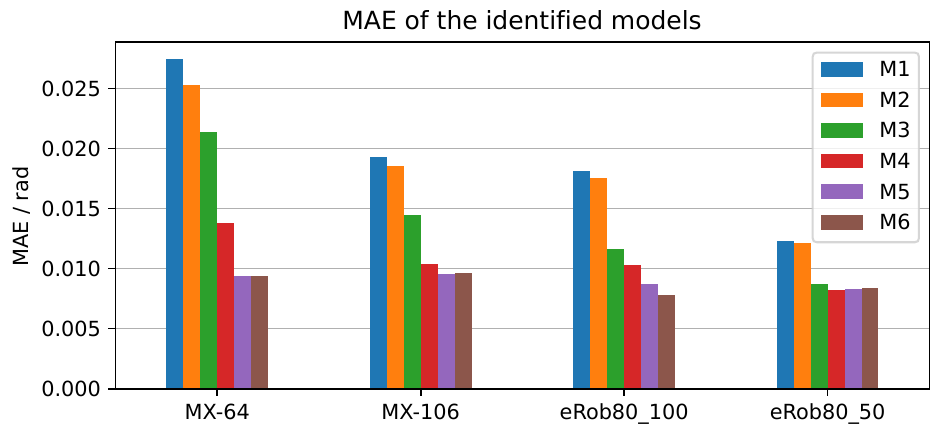}
    \caption{
        MAE obtained on the validation logs after identification for each model on the four servo actuators.
    }
    \label{fig:identification}
\end{figure}

The MAE obtained on the validation logs after identification is presented in Fig. \ref{fig:identification}.
The results show that, for the Dynamixel servo actuators, there are consistent improvements in MAE from models $\mathcal{M}_1$ through $\mathcal{M}_5$ 
and no improvement for the $\mathcal{M}_6$ model. For the eRob80:100 the best performance is 
achieved with the $\mathcal{M}_6$ model, showing gradual improvements from $\mathcal{M}_1$ to $\mathcal{M}_6$. In contrast, for the eRob80:50, there 
is no clear improvement beyond the $\mathcal{M}_3$ model.

Overall, the best model for each servo actuator significantly improves the MAE compared to the Coulomb-Viscous model. 
Specifically, the MAE is reduced by a factor of 2.93 for the MX-64 servo actuator, 2.02 for the MX-106, 2.34 for 
the eRob80:100, and 1.51 for the eRob80:50.

\subsection{Validation on 2R arms}

\begin{figure}[!ht]
    \centering
    \par
    \vspace{1em}
    \includegraphics[width=.9\columnwidth]{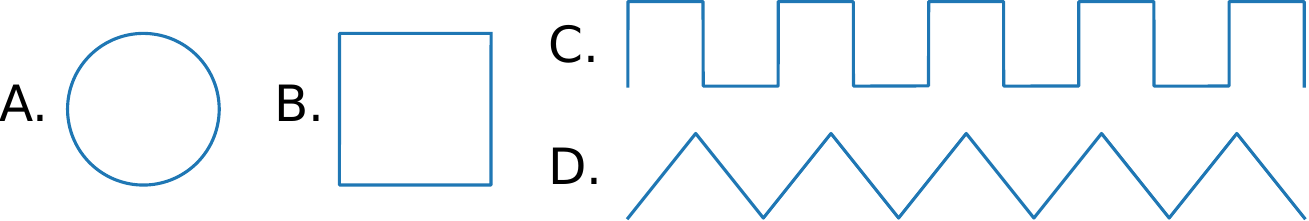}
    \caption{
        Paths recorded and simulated on the 2R arms: \\Circle (\textbf{A.}), Square (\textbf{B.}), Square wave (\textbf{C.}), Triangular wave (\textbf{D.}).
    }
    \label{fig:rr_traj}
\end{figure}

To demonstrate the improvement brought by the extended friction models on more complex systems, 
two 2R arms are used. The desired test paths are presented in Fig. \ref{fig:rr_traj} and executed both on the real robots and in simulation.
The first arm is composed of Dynamixel MX-106 and MX-64, and the second one of 
eRob80:100 and eRob80:50. The arms are controlled with the same modality as during identification.
The simulation is performed using MuJoCo, with the friction models implemented as described in section \ref{subsec:sim-engine}.

Each trajectory is recorded and simulated using both high and low proportional gains. 
High gains provide precise control, while low gains offer loose control, as typically 
used in RL. Fig. \ref{fig:catch-eye} details a log obtained with the Dynamixel arm on a triangular wave
with low gains.

\begin{figure}[!ht]
    \includegraphics[width=1.\columnwidth]{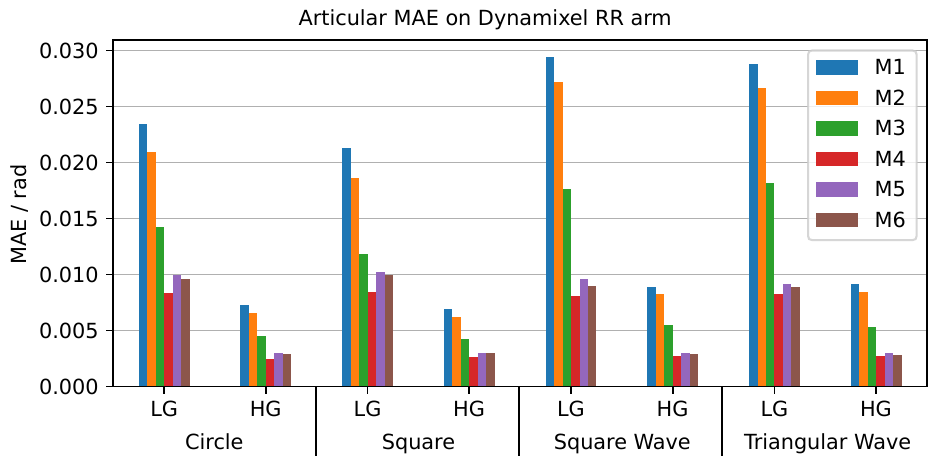}
    \includegraphics[width=1.\columnwidth]{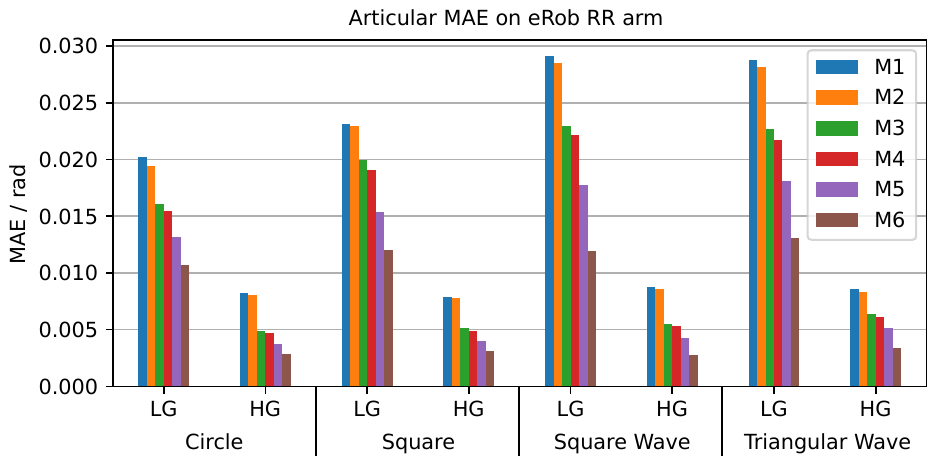}
    \caption{
        MAE obtained on the 2R arms trajectories for each model. 
        Each trajectory is recorded using Low Gains (LG) and High Gains (HG) for the controller.
    }
    \label{fig:rr_results}
\end{figure}

Fig. \ref{fig:rr_results} presents the MAE for each model on the 2R arm trajectories. 
The results show that the $\mathcal{M}_4$ model outperforms others for the Dynamixel 2R arm, with an 
MAE over twice as low as the Coulomb-Viscous model across all trajectory and gain configurations. 
Although the $\mathcal{M}_5$ model performs better during the identification phase for Dynamixel servo actuators, 
it does not improve results on the 2R arm, likely due to overfitting. For the eRob80 2R arm, 
the $\mathcal{M}_6$ model is the most accurate, also achieving an MAE more than twice as low as the Coulomb-Viscous model.

\section{Conclusion}

This work presents a method for simulating servo actuators using extended friction models. 
The proposed method and models demonstrate their effectiveness, significantly improving 
the simulation of four distinct servo actuators. An open-source implementation of the proposed models and 
the identification method is available at \href{https://github.com/Rhoban/bam}{https://github.com/Rhoban/bam}.

\subsection{Limitations}

The temperature of the servo actuators is not considered in this work~\cite{bittencourt2012static}. 
Accounting for this would require the implementation of a thermal model 
in the servo actuator simulation, along with methods to measure and control the temperature during data logging.
Radial forces are also neglected in the models to limit the complexity of the experimental setup.
Finally the dwell time~\cite{gonthier2004regularized}, which refers to the period during which two surfaces in contact 
remain stationary relative to each other before sliding resumes, is not considered. 
This effect could be modeled by introducing a delay in the friction response.

\subsection{Future Works}

Future works will focus on the implementation of the proposed models in a physics engine, aiming 
to simulate more accurately humanoids and other complex systems. The use of the proposed models
should allow to improve the transferability of reinforcement learning policies from simulation to real-world.

\bibliographystyle{ieeetr}
\bibliography{biblio}

\begin{thebibliography}{10}

\bibitem{rlRoboticsSurvey2013}
J.~Kober, J.~A. Bagnell, and J.~Peters, ``Reinforcement learning in robotics:
  {A} survey,'' {\em Int. J. Robotics Res.}, vol.~32, no.~11, pp.~1238--1274,
  2013.

\bibitem{todorov2012mujoco}
E.~Todorov, T.~Erez, and Y.~Tassa, ``Mujoco: A physics engine for model-based
  control,'' in {\em IEEE/RSJ International Conference on Intelligent Robots
  and Systems (IROS)}, pp.~5026--5033, 2012.

\bibitem{isaacGym}
V.~Makoviychuk, L.~Wawrzyniak, {\em et~al.}, ``Isaac gym: High performance
  {GPU} based physics simulation for robot learning,'' in {\em NeurIPS Datasets
  and Benchmarks} (J.~Vanschoren and S.~Yeung, eds.), 2021.

\bibitem{zhu2021survey}
W.~Zhu, X.~Guo, {\em et~al.}, ``A survey of sim-to-real transfer techniques
  applied to reinforcement learning for bioinspired robots,'' {\em IEEE
  Transactions on Neural Networks and Learning Systems}, vol.~34, no.~7,
  pp.~3444--3459, 2021.

\bibitem{openAI2020dexterous}
M.~Andrychowicz, B.~Baker, {\em et~al.}, ``Learning dexterous in-hand
  manipulation,'' {\em Int. J. Robotics Res.}, vol.~39, no.~1, 2020.

\bibitem{fabre2017dynaban}
R.~Fabre, Q.~Rouxel, G.~Passault, S.~N’Guyen, and O.~Ly, ``Dynaban, an
  open-source alternative firmware for dynamixel servo-motors,'' in {\em
  RoboCup 2016: Robot World Cup XX 20}, pp.~169--177, Springer, 2017.

\bibitem{masuda2023sim}
S.~Masuda and K.~Takahashi, ``Sim-to-real transfer of compliant bipedal
  locomotion on torque sensor-less gear-driven humanoid,'' in {\em 2023
  IEEE-RAS 22nd International Conference on Humanoid Robots (Humanoids)},
  pp.~1--8, IEEE, 2023.

\bibitem{actuatorNet2019}
J.~Lee, A.~Dosovitskiy, {\em et~al.}, ``Learning agile and dynamic motor skills
  for legged robots,'' {\em Sci. Robotics}, vol.~4, no.~26, 2019.

\bibitem{transformerDisney2023}
A.~Serifi, E.~Knoop, {\em et~al.}, ``Transformer-based neural augmentation of
  robot simulation representations,'' {\em IEEE Robotics Autom. Lett.}, vol.~8,
  no.~6, pp.~3748--3755, 2023.

\bibitem{albender2008characterization}
F.~Al-Bender and J.~Swevers, ``Characterization of friction force dynamics,''
  {\em IEEE Control Systems Magazine}, vol.~28, no.~6, pp.~64--81, 2008.

\bibitem{coulomb1785theorie}
C.~A. Coulomb, {\em Th{\'e}orie des machines simples: en ayant {\'e}gard au
  frottement de leurs parties, et {\`a} la roideur de cordages}.
\newblock 1785.

\bibitem{pennestri2016review}
E.~Pennestr{\`\i}, V.~Rossi, {\em et~al.}, ``Review and comparison of dry
  friction force models,'' {\em Nonlinear dynamics}, vol.~83, pp.~1785--1801,
  2016.

\bibitem{bittencourt2012static}
A.~C. Bittencourt and S.~Gunnarsson, ``Static friction in a robot
  joint—modeling and identification of load and temperature effects,'' {\em
  ASME J. Dyn. Sys., Meas., Control}, vol.~134, 2012.

\bibitem{mx64}
{Robotis Co.,Ltd.}, ``Dynamixel mx-64.''
  \url{https://emanual.robotis.com/docs/en/dxl/mx/mx-64/}, 2024.
\newblock Accessed: 2024-08-28.

\bibitem{mx106}
{Robotis Co.,Ltd.}, ``Dynamixel mx-106.''
  \url{https://emanual.robotis.com/docs/en/dxl/mx/mx-106-2/}, 2024.
\newblock Accessed: 2024-08-28.

\bibitem{erob80}
{ZeroErr Control Co.,Ltd}, ``erob-80.''
  \url{https://en.zeroerr.cn/rotary_actuators/erob80i}, 2024.
\newblock Accessed: 2024-08-28.

\bibitem{lynch2017modern}
K.~M. Lynch and F.~C. Park, {\em Modern Robotics}.
\newblock Cambridge University Press, 2017.

\bibitem{wilfrido2021load}
P.~Q.~C. Wilfrido, A.~Gabriel, {\em et~al.}, ``Load-dependent friction laws of
  three models of harmonic drive gearboxes identified by using a force transfer
  diagram,'' in {\em International Conference on Mechanical and Aerospace
  Engineering (ICMAE)}, pp.~239--244, IEEE, 2021.

\bibitem{zhu2019design}
T.~Zhu, J.~Hooks, and D.~Hong, ``Design, modeling, and analysis of a liquid
  cooled proprioceptive actuator for legged robots,'' in {\em 2019 IEEE/ASME
  International Conference on Advanced Intelligent Mechatronics (AIM)},
  pp.~36--43, IEEE, 2019.

\bibitem{mori2023identification}
K.~Mori and G.~Venture, ``Identification of the gear transmission’s
  efficiency by neural network,'' in {\em IFToMM World Congress on Mechanism
  and Machine Science}, pp.~899--909, Springer, 2023.

\bibitem{wang2015directional}
A.~Wang and S.~Kim, ``Directional efficiency in geared transmissions:
  Characterization of backdrivability towards improved proprioceptive
  control,'' in {\em IEEE International Conference on Robotics and Automation
  (ICRA)}, pp.~1055--1062, 2015.

\bibitem{sihite2019derivation}
E.~N. Sihite, D.~J. Yang, and T.~R. Bewley, ``Derivation of a new drive/coast
  motor driver model for real-time brushed dc motor control, and validation on
  a mip robot,'' in {\em IEEE International Conference on Automation Science
  and Engineering (CASE)}, pp.~1099--1105, 2019.

\bibitem{hansen2001completelyCMAES}
N.~Hansen and A.~Ostermeier, ``Completely derandomized self-adaptation in
  evolution strategies,'' {\em Evol. Comput.}, vol.~9, no.~2, pp.~159--195,
  2001.

\bibitem{akiba2019optuna}
T.~Akiba, S.~Sano, {\em et~al.}, ``Optuna: A next-generation hyperparameter
  optimization framework,'' in {\em International Conference on Knowledge
  Discovery \& Data Mining (ACM SIGKDD)}, pp.~2623--2631, 2019.

\bibitem{gonthier2004regularized}
Y.~Gonthier, J.~McPhee, {\em et~al.}, ``A regularized contact model with
  asymmetric damping and dwell-time dependent friction,'' {\em Multibody System
  Dynamics}, vol.~11, pp.~209--233, 2004.

\end{thebibliography}

\end{document}